%% file: paper.tex
\documentclass[letterpaper, 10 pt, conference]{ieeeconf}  % Comment this line out if you need a4paper

\IEEEoverridecommandlockouts                              % This command is only needed if 
\usepackage{graphics} % for pdf, bitmapped graphics files
\usepackage{epsfig} % for postscript graphics files
\usepackage{mathptmx} % assumes new font selection scheme installed
\usepackage{times} % assumes new font selection scheme installed
\usepackage{amsmath} % assumes amsmath package installed
\usepackage{amssymb}  % assumes amsmath package installed
\usepackage{color}
\usepackage{subfig}
\usepackage{cite}
\usepackage{graphicx}
\usepackage{siunitx}
\usepackage{soul}
\usepackage{tikz}
\usepackage{booktabs}
\usepackage{xspace}
\usepackage{url}

\graphicspath{{figs/}}

\title{\LARGE \bf
Fast and Accurate: Video Enhancement Using Sparse Depth}

\author{Yu Feng$^{1}$, Patrick Hansen$^{2}$, Paul N. Whatmough$^{2}$, Guoyu Lu$^{3}$, and Yuhao Zhu$^{1}$% <-this % stops a space
\thanks{$^{1}$These authors are with the Department of Computer Science, University of Rochester, Rochester, NY, USA. {\tt\small yfeng28@ur.rochester.edu}, {\tt\small yzhu@rochester.edu}}%
\thanks{$^{2}$These authors are with Arm Research, Boston, MA, USA. {\tt\small patrick.hansen@arm.com}, {\tt\small paul.whatmough@arm.com}}
\thanks{$^{3}$This author is with Rochester Institute of Technology, Rochester, NY, USA. {\tt\small luguoyu@cis.rit.edu}}
}

\begin{document}

\maketitle
\thispagestyle{empty}
\pagestyle{empty}

%%%%%%%%%%%%%%%%%%%%%%%%%%%%%%%%%%%%%%%%%%%%%%%%%%%%%%%%%%%%%%%%%%%%%%%%%%%%%%%%

\input{macro}
\input{abstract}

%%%%%%%%%%%%%%%%%%%%%%%%%%%%%%%%%%%%%%%%%%%%%%%%%%%%%%%%%%%%%%%%%%%%%%%%%%%%%%%%
% main paper

\input{intro}

\input{related}

\input{framework}

\input{setup}

\input{eval}

\input{conc}

%%%%%%%%%%%%%%%%%%%%%%%%%%%%%%%%%%%%%%%%%%%%%%%%%%%%%%%%%%%%%%%%%%%%%%%%%%%%%%%%

%%%%%%%%%%%%%%%%%%%%%%%%%%%%%%%%%%%%%%%%%%%%%%%%%%%%%%%%%%%%%%%%%%%%%%%%%%%%%%%%

%bibliography
\bibliographystyle{IEEEtranS}
\bibliography{refs}

\end{document}

%% file: macro.tex
\newcommand{\website}[1]{{\tt #1}}
\newcommand{\program}[1]{{\tt #1}}
\newcommand{\benchmark}[1]{{\it #1}}
\newcommand{\fixme}[1]{{\textcolor{red}{\textit{#1}}}}

\newcommand*\circled[2]{\tikz[baseline=(char.base)]{
            \node[shape=circle,fill=black,inner sep=1pt] (char) {\textcolor{#1}{{\footnotesize #2}}};}}

\ifx\figurename\undefined \def\figurename{Figure}\fi
\renewcommand{\figurename}{Fig.}
\renewcommand{\paragraph}[1]{\textbf{#1}~~}
\newcommand{\figline}{{\vspace*{.05in}\hline}}

\newcommand{\Alg}[1]{Alg.~\ref{#1}}
\newcommand{\Sect}[1]{Sec.~\ref{#1}}
\newcommand{\Fig}[1]{Fig.~\ref{#1}}
\newcommand{\Tbl}[1]{Tbl.~\ref{#1}}
\newcommand{\Equ}[1]{Eqn.~\ref{#1}}
\newcommand{\Apx}[1]{Appendix~\ref{#1}}

\newcommand{\specialcell}[2][c]{\begin{tabular}[#1]{@{}l@{}}#2\end{tabular}}
\newcommand{\note}[1]{\textcolor{red}{#1}}

\newcommand{\greenweb}{{\fontfamily{cmtt}\selectfont GreenWeb}\xspace}
\newcommand{\autogreen}{\textsc{AutoGreen}\xspace}
\newcommand{\proj}{\textsc{ASV}\xspace}
\newcommand{\mode}[1]{\underline{\textsc{#1}}\xspace}

\newcommand{\floor}[1]{\left\lfloor #1 \right\rfloor}
\newcommand{\ceil}[1]{\left\lceil #1 \right\rceil}

\newcommand{\RNum}[1]{\uppercase\expandafter{\romannumeral #1\relax}}

% checkmark and xmark in the pifont package
%\newcommand{\cmark}{\ding{51}}
%\newcommand{\xmark}{\ding{55}}

%% file: abstract.tex
\begin{abstract}

This paper presents a general framework to build fast and accurate algorithms for video enhancement tasks such as super-resolution, deblurring, and denoising. Essential to our framework is the realization that the accuracy, rather than the density, of pixel flows is what is required for high-quality video enhancement. Most of prior works take the opposite approach: they estimate dense (per-pixel)---but generally less robust---flows, mostly using computationally costly algorithms.
Instead, we propose a lightweight flow estimation algorithm; it fuses the sparse point cloud data and (even sparser and less reliable) IMU data available in modern autonomous agents to estimate the flow information. Building on top of the flow estimation, we demonstrate a general framework that integrates the flows in a plug-and-play fashion with different task-specific layers. Algorithms built in our framework achieve 1.78$\times$ --- 187.41$\times$ speedup while providing a 0.42dB -- 6.70 dB quality improvement over competing methods.
 
\end{abstract}

%% file: intro.tex
\section{Introduction}
\label{sec:intro}

Video enhancement tasks ranging from super resolution~\cite{wang2019edvr,caballero2017vespcn,tian2020tdan,haris2019rbpn}, deblurring~\cite{sim2019deep_deblur,zhong2020estrnn,Nah_2017_deblur}, and denoising~\cite{tassano2019dvdnet,claus2019denoise} are becoming increasingly important for intelligent systems such as smartphones and Augmented Reality (AR) glasses. High-quality videos are also critical to various robotics tasks, such as SLAM \cite{roznere2019mutual, cho2017visibility}, visual odometry \cite{gomez2018learning}, object detection \cite{parekh2021survey}, and surveillance \cite{rao2010image}.

%Video enhancement typically relies on extracting temporal correlations across adjacent frames. Temporal correlation is obtained through either \textit{explicit} motion estimation algorithms such as optical flow~\cite{tassano2019dvdnet, caballero2017vespcn, haris2019rbpn} or through \textit{implicit} methods that extract feature correlations~\cite{tian2020tdan, wang2019edvr, zhong2020estrnn}.

Video enhancement systems today face a fundamental dilemma. High quality enhancement benefits from accurately extracting temporal flows across adjacent frames, which, however, is difficult to obtain from low-quality videos (e.g., low-resolution, noisy). As a result, video enhancement usually requires expensive optical flow algorithms, usually in the form of Deep Neural Networks (DNNs), to extract dense flows, leading to a low execution speed. As video enhancement tasks execute on resource-limited mobile devices and potentially in real time, there is a need for high-speed and high-quality video enhancement.

We propose a method to simultaneously increase the quality and the execution speed of video enhancement tasks. Our work is based on the realization that the accuracy, rather than the density, of the flow estimation is what high-quality enhancement requires. We propose an algorithm to estimate accurate, but sparse, flows using LiDAR-generated point clouds. Coupled with the flow estimation algorithm, we demonstrate a generic framework that incorporates the flows to build video enhancement DNNs, which are lightweight by design owing to the assistance of accurate flows.

Our flow estimation is accurate because it does not rely on the image content, which is necessarily of low-quality in video enhancement tasks. Instead, we generate flows using the accurate depth information from LiDAR point cloud assisted with the less reliable IMU information. By exploiting the spatial geometry of scene depth and the agent's rough ego-motion (through IMU), our algorithm estimates the flows in videos using a purely analytical approach without complex feature extraction, matching, optimization, and learning used in conventional flow estimation algorithms.

%We use LiDAR because it is readily available in robots, autonomous vehicles, and even smartphones (e.g., iPhone 12 Pro), and it provides accurate depth information.

%A key challenge we address is that the motion obtained from LiDAR-generated point cloud is typically sparse, because today's LiDAR resolution is at least an order of magnitude lower than that of cameras. We propose optimizations that increase the point cloud density and, by extension, the motion density, further improving the task quality.

Building on top of the lightweight flow estimation, we demonstrate a general framework that integrates the flows for video enhancement. The framework consists of a common temporal alignment front-end and a task-specific back-end. The front-end temporally aligns a sequence of frames by warping and concatenating frames using the estimated flows; the back-end extracts task-specific features to synthesize high-quality videos. Different from prior works that specialize the temporal alignment module for a specific task, our unified temporal alignment module broadly applies to different enhancement tasks and, thus, empowers algorithm developers to focus energy on the task-specific back-end.

%, both the implicit and explicit ones. Explicit methods use explicit, usually heavy-duty, motion estimation algorithms such as dense optical flow~\cite{tassano2019dvdnet, caballero2017vespcn, haris2019rbpn}. We replace their motion estimation modules directly with our motion estimation algorithm, which is both faster and more accurate.

%Implicit methods extract temporal correlation in the feature space~\cite{tian2020tdan, wang2019edvr, zhong2020estrnn}. We use our motions to warp adjacent frames to the target (to-be-enhanced) frame, and replace the original input frames with the warped frames. This strategy allows us provide an initial alignment of frames, which simplifies the rest of the system. Such simplifications include reducing the required receptive field and/or the complexity of the image reconstruction network.

We demonstrate our framework on a range of video enhancement tasks including super resolution, deblurring, and denoising on the widely-used KITTI dataset~\cite{Geiger2013kitti}. Across all tasks, our system has better enhancement quality than state-of-the-art algorithms measured in common metrics such as Peak Signal-to-Noise Ratio (PSNR) and Structure Similarity Index Measure (SSIM)~\cite{wang2004image}. Meanwhile, we improve the execution speed on all tasks by a factor of 8.4 on average (up to 187.4 
times). The code will be open-sourced.

%Our main contributions of the paper are as follows:
%
%\begin{itemize}
%	\item We propose a LiDAR-guided motion estimation algorithm is extremely lightweight and accurate.
%\end{itemize}

%% file: related.tex
\section{Related Work}
\label{sec:related}

%\begin{figure}[t]
%\centering
%\subfloat[General structure of explicit algorithms.]{
%	\label{fig:image_warp}	
%	\includegraphics[width=.95\columnwidth]{warp_input}}
%\vspace{5pt}
%\vspace{-5pt}
%\subfloat[General structure of implicit algorithms.]{
%	\label{fig:implicit_alignment}
%	\includegraphics[width=.95\columnwidth]{implicit_alignment}}
%\vspace{5pt}
%\caption{General structures of representative video enhancement algorithms. They rely on extracting temporal correlation, through explicit or implicit motion estimation, across adjacent frames. Our proposed approach improves the speed and quality of both explicit and implicit methods by providing accurate and fast motion estimation.}
%%\vspace{-15pt}
%\label{fig:alignment}
%\end{figure}

\paragraph{Video Enhancement} The general theme in today's video enhancement algorithms is to first align neighboring frames from time $t-n$ to time $t+m$ and then fuse the aligned frames to enhance the target frame $t$. Much of the prior innovations lie in how to better align frames.

Alignment could be done explicitly or implicitly. Explicit approaches perform an explicit flow estimation between frames~\cite{caballero2017vespcn, tassano2019dvdnet, haris2019rbpn}. The flows are then used to align frames either in the image space~\cite{caballero2017vespcn, haris2019rbpn} or in the feature space~\cite{tassano2019dvdnet}. Obtaining accurate flows typically requires expensive flow estimation algorithms (e.g., dense optical flow~\cite{haris2019rbpn} or complicated DNNs~\cite{tassano2019dvdnet,caballero2017vespcn}), which lead to low execution speed.
%\Fig{fig:image_warp} shows a typical structure of the explicit approach, and is representative of algorithms such as RBPN~\cite{haris2019rbpn} and VESPCN~\cite{caballero2017vespcn}.
Implicit approaches, instead, align frames in latent space using algorithms such as deformable convolution~\cite{tian2020tdan,wang2019edvr} or recurrent neural networks~\cite{zhong2020estrnn}. Classic examples include EDVR~\cite{wang2019edvr}, TDAN~\cite{tian2020tdan} and ESTRNN~\cite{zhong2020estrnn}. These algorithms tend to be more accurate than explicit approaches when the temporal correlation is not obvious in pixels pace.
%\Fig{fig:implicit_alignment} shows the overall architecture.

%Motion compensation plays an important role in many video enhancement tasks, such as super-resolution~\cite{}, video denoising~\cite{} and even video compression~\cite{}. Prior works use motion compensation to align the temporal information and better leverage the temporal information~\cite{}. Most studies use the off-the-shelf motion estimation algorithms to pre-process the inputs~\cite{} or warp the latent feature space~\cite{}. Recent studies also incorporate the motion estimation into their DNN algorithms and co-train the network jointly.

Our work differs from prior works in two main ways. First, both implicit and explicit approaches are computationally-heavy, as they extract flows from purely the vision modality. We demonstrate a very fast algorithm to extra flows by fusing LiDAR and IMU data. We show that accurate flows enable a simple downstream DNN design, achieving state-of-the-art task quality while being an order of magnitude faster. Second, the alignment modules in prior works usually are specialized for specific enhancement tasks. We instead show a common alignment module based on our estimated flows broadly applies to a range of video enhancement tasks. This greatly eases development and deployment effort in practice.

\paragraph{LiDAR-Guided Vision} Fusing point clouds and images is known to improve the quality of vision tasks such as object detection~\cite{chen2017multi, xu2018pointfusion, yoo20203d}, segmentation~\cite{el2019rgb, meyer2019sensor}, and stereo matching~\cite{cheng2019noise, wang20193d}, but literature is scarce in LiDAR-camera fusion for video enhancement.

Fusion networks usually extract features from (LiDAR-generated) point clouds and images, and align/fuse the two sets of features before feeding them to the task-specific block. Unlike prior fusion algorithms that extract features from point clouds, we propose a different way of using point cloud data, i.e., estimating explicit pixel flow from point clouds. The estimated flows are accurate and, thus, provide targeted guidance to video enhancement tasks.

\paragraph{Flow Estimation} Estimating flows between frames is a fundamental building block. Video-based flow estimation has made great strides through DNNs~\cite{FlownetC2015, ranjan2017optical, sun2018pwc}. These methods, however, are computationally intensive. When incorporated into a high-level vision task such as deblurring and denoising, the flow estimation quickly becomes a speed bottleneck. Many flow estimations algorithms use only video frames, which, while is less restrictive, also means the flow accuracy degrades when operating on low-quality videos. Our method is image content-independent and thus better estimates flows from low-quality videos. It is also very fast, because it relies purely on simply geometric transformations.

Existing video enhancement tasks usually use dense and per-pixel flow estimation~\cite{haris2019rbpn, caballero2017vespcn, tassano2019dvdnet}. In contrast, our method generates sparse flows from point clouds. A key contribution of our work is to demonstrate that even a sparse flow can greatly boost the quality of video enhancement.

%% file: framework.tex
\section{Main Idea and Optimizations}
\label{sec:design}

\begin{figure}[t]
\centering
\includegraphics[trim=0 0 0 0, clip, width=\columnwidth]{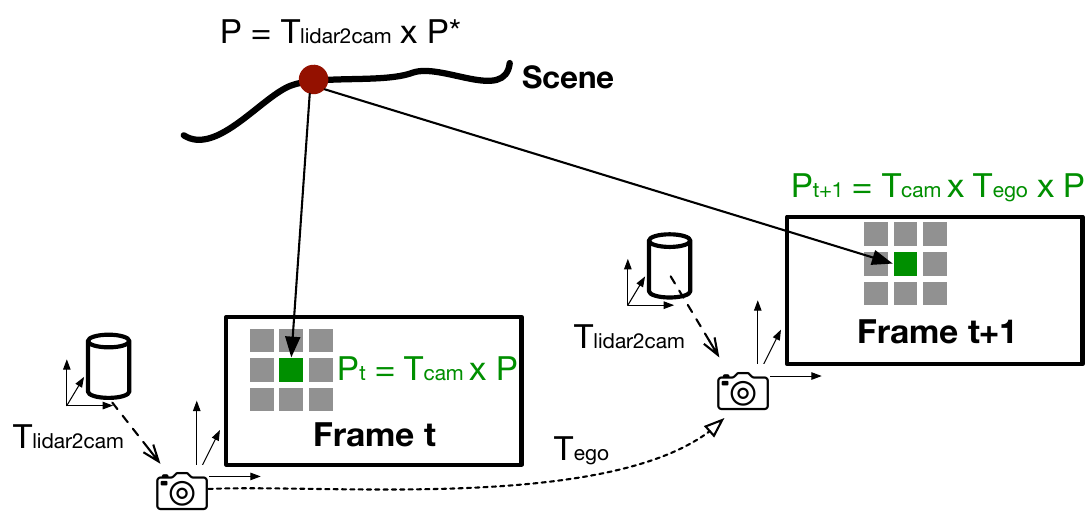}
\caption{LiDAR-guided flow estimation. $P^*$ is the 3D coordinates of a point in the LiDAR coordinate system at time $t$. $T_{lidar2cam}$ is the transformation matrix from the LiDAR coordinate system to the camera coordinate system, which is fixed over time assuming the configuration of the LiDAR and camera is rigid. $T_{cam}$ is the camera matrix. $T_{ego}$ is the camera egomotion from Frame $t$ to Frame $t+1$.}
\label{fig:motion}
\end{figure}

\begin{figure*}[t]
\centering
\includegraphics[trim=0 0 0 0, clip, width=1.8\columnwidth]{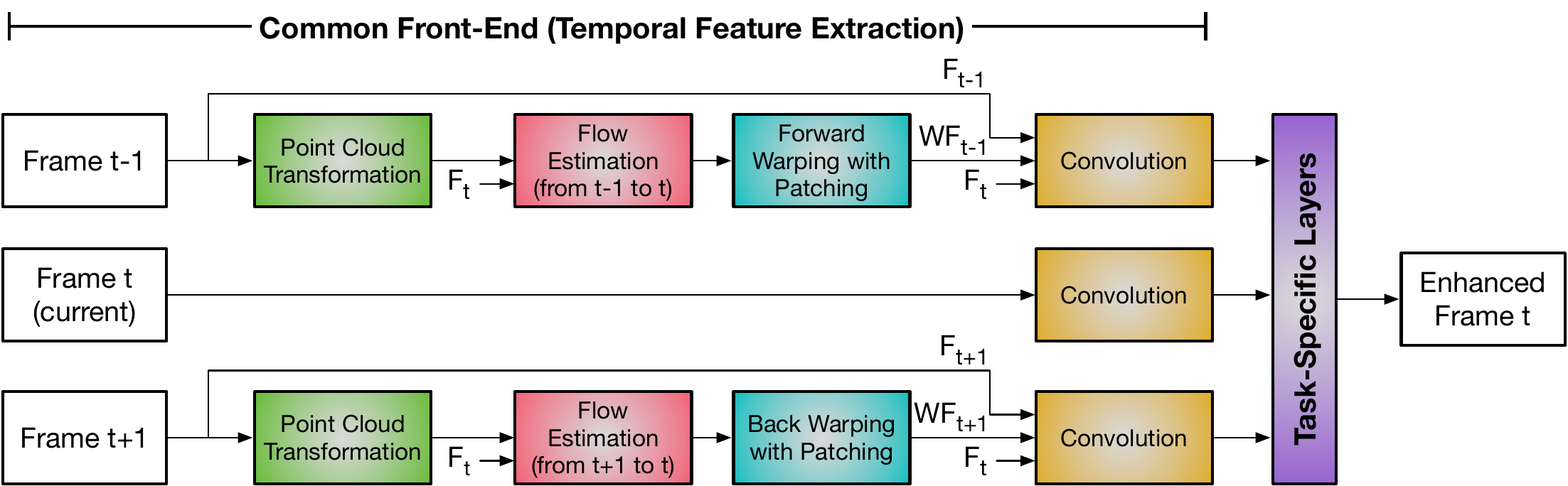}
\caption{Overview of our two-stage video enhancement DNN architecture. The front-end performs lightweight flow estimation to align (previous and later) frames  with the current frame at time \textit{t} in order to extract temporal features. The extracted features carry temporal correlations across frames and are then processed by task-specific layers to produce an enhanced frame. We merge point clouds (using the estimated ego-motion) before flow estimation and warp pixels in patches after flow estimation, both to mitigate the sparsity of LiDAR-generated point clouds.}
\label{fig:framework}
\end{figure*}

We first describe the lightweight flow estimation algorithm (\Sect{sec:design:motion}), followed by a generic DNN architecture that integrates the flows for video enhancement (\Sect{sec:design:idea}).

\subsection{Lightweight and Accurate Flow Estimation}
\label{sec:design:motion}

\paragraph{Overall Algorithm} The key idea is to use the depth data from LiDAR to generate flows in a lightweight fashion. \Fig{fig:motion} illustrates the idea. For any point $P^*$ in a point cloud, it is captured by two consecutive camera frames. At time $t$, $P^*$'s coordinates in the camera coordinate system are $P = T_{lidar2cam} \times P^* $, where $T_{lidar2cam}$ is the LiDAR to camera transformation matrix, which is usually pre-calibrated. Thus, the corresponding pixel coordinates in the image at time $t$ are $P_t = T_{cam} \times P$, where $T_{cam}$ is the camera matrix.

At time $t+1$, the coordinates of the same point in the scene in the camera coordinate system are $T_{ego} \times P$, where $T_{ego}$ is the transformation matrix of the camera egomotion. Thus, the pixel coordinates of the point at $t+1$ are $P_{t+1} = T_{cam} \times T_{ego} \times P$. Accordingly, the pixel's motion vector can be calculated in a computationally very lightweight manner:

\begin{align}
  \delta_t =& P_{t+1} - P_t \nonumber \\
           =& T_{cam} \times T_{ego} \times T_{lidar2cam} \times P^* - T_{cam}\times T_{lidar2cam} \times P^*.
  \label{eq:motion}
\end{align}

\paragraph{Egomotion} The camera egomotion $T_{ego}$ could be derived in a range of different methods. In our system, we estimate $T_{ego}$ using the measurements from the IMU, which is widely available in virtually all intelligent devices. We note that the IMU data, while being a readily available sensor modality, is known to be a rough and imprecise estimation of the true egomotion~\cite{cho2018evaluation}. One of our contributions is to show how the rough egomotion estimation can provide decent flow estimation for high-quality video enhancement.

The IMU provides the translational acceleration ($\mathbf{\hat a}$) and the angular velocity ($\mathbf{\hat \omega}$). Given $\mathbf{\hat a}$, the translation component $T_{3\times 1}$ in $T_{ego}$ is calculated by:

\begin{align}
    T_{3\times 1} = \begin{bmatrix}
                    {\overline{\Delta x}} &
                    {\overline{\Delta y}} &
                    {\overline{\Delta z}} 
                    \end{bmatrix}
    \label{equ:tran}
\end{align}

\noindent where $\overline{\Delta x}$, $\overline{\Delta y}$, and $\overline{\Delta z}$ are the three translational displacements integrated from $\mathbf{\hat a}$ using Euler's method. Similarly, the rotational component $R_{3\times 3}$ in $T_{ego}$ is estimated from $\mathbf{\hat \omega}$ as:

\begin{align}
  R_{3\times 3} = R^{y}_{3\times 3} \times R^{p}_{3\times 3} \times R^{r}_{3\times 3}
%    R_{3\times 3} & = \begin{bmatrix}
%    cos\large({\overline{\Delta \alpha}}\large) & sin\large({\overline{\Delta \alpha}}\large) & 0\\
%    -sin\large({\overline{\Delta \alpha}}\large) & 
%    cos\large({\overline{\Delta \alpha}}\large) & 0\\
%    0 & 0 & 1
%    \end{bmatrix} \nonumber \\ 
%    & \times \begin{bmatrix}
%    cos\large({\overline{\Delta \beta}}\large) & 0 &  -sin\large({\overline{\Delta \beta}}\large)\\
%     0 & 1 & 0 \\
%    sin\large({\overline{\Delta \beta}}\large) & 0 &  cos\large({\overline{\Delta \beta}}\large)
%    \end{bmatrix} \nonumber \\
%    & \times \begin{bmatrix}
%    1 & 0 & 0 \\
%    0 & cos\large({\overline{\Delta \gamma}}\large) & sin\large({\overline{\Delta \gamma}}\large) \\
%    0 & -sin\large({\overline{\Delta \gamma}}\large) & cos\large({\overline{\Delta \gamma}}\large)
%    \end{bmatrix}
    \label{equ:rot}
\end{align}

\noindent where $R^{y}_{3\times 3}$, $R^{p}_{3\times 3}$, and $R^{r}_{3\times 3}$ denote the three rotational matrices, which are integrated from the three rotational displacements in $\mathbf{\hat \omega}$ using Euler's method.

% $\overline{\Delta \alpha}$, $\overline{\Delta \beta}$, and $\overline{\Delta \gamma}$ are .

%Alternatively, $T_{ego}$ could be estimated through point cloud registration~\cite{besl1992method, zhang1994iterative}, which is generally more costly but more accurate. As we will demonstrate later (\Sect{sec:eval:quality}), even with relatively less accurate estimations from the IMU, our method can still reliably deliver higher task accuracy while greatly reducing the execution time. When given more accurate egomotions (e.g., from point cloud registration), our framework would have a higher video enhancement quality.

A key reason why video enhancement benefits from our flow estimation is that our algorithm is purely based on 3D geometry and geometric transformation \textit{without} relying on the image content. No pixel content participates in the flow estimation \Equ{eq:motion}. Therefore, it estimates flows accurately even when the image content is of low-quality, e.g., low resolution or noisy, which is exactly the kind of scenario video enhancement tasks target at.

\subsection{A Generic DNN Architecture}
\label{sec:design:idea}

We see our flow estimation as a building block for simultaneously improving the quality and execution speed of video enhancement. To that end, we propose a generic DNN architecture that incorporates the estimated flows for a range of video enhancement tasks. \Fig{fig:framework} shows an overview of the architecture, which consists of two main modules: a common frame fusion front-end and a task-specific back-end.

\begin{figure}[t]
\centering
\includegraphics[trim=0 0 0 0, clip, width=1\columnwidth]{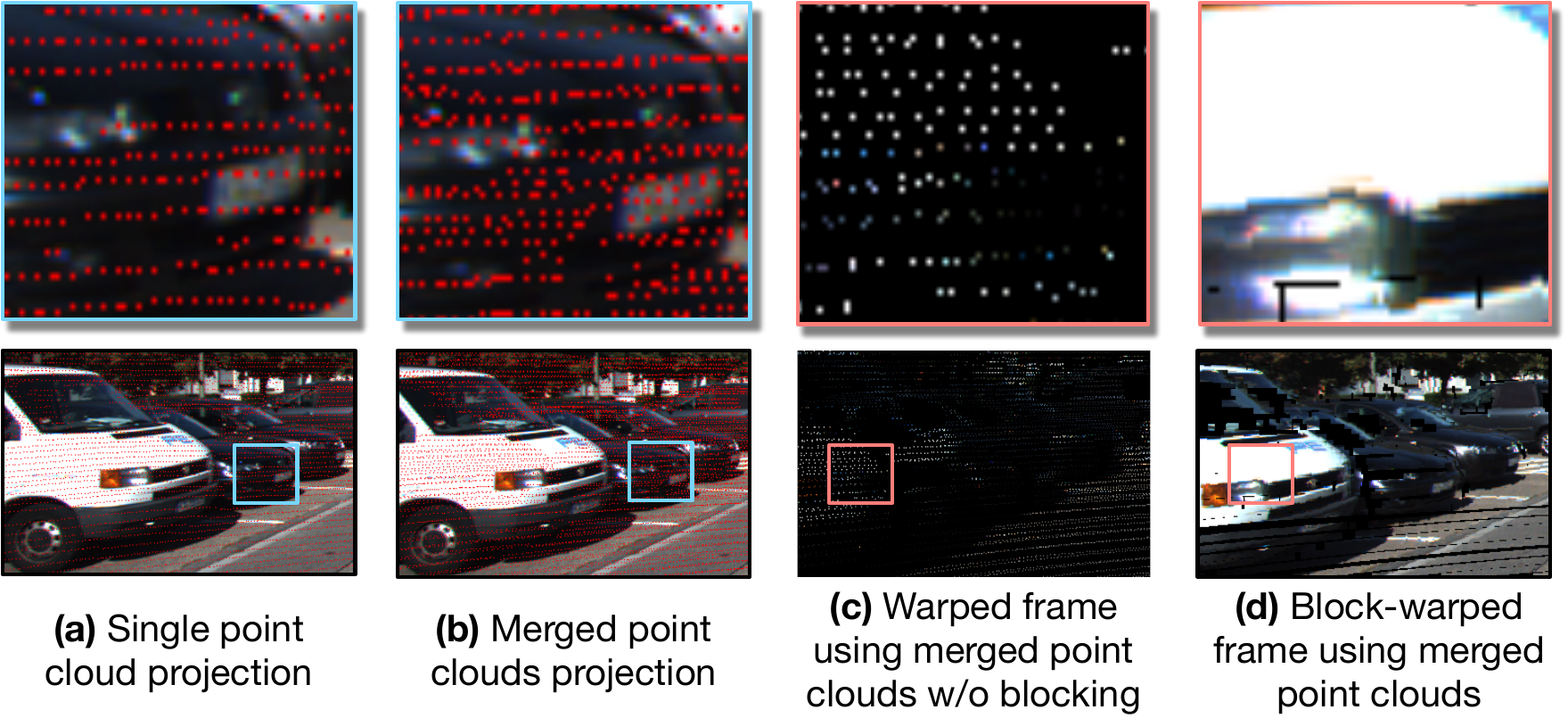}
\caption{Effectiveness of techniques to increase point cloud density. (a): frame overlaid with a projected single point cloud (red pixels are projected points). (b): frame overlaid with the projection of five merged point clouds. (c): warped frame using flows estimated from five merged point clouds without blocking. (d): frame warped using flows from five merged point clouds and warped in $5\times 5$ blocks.}
\label{fig:comp}
\end{figure}

\paragraph{Temporal Feature Extraction} Our network uses a common front-end shared across different enhancement tasks. The goal of the front-end is to extract temporal correlations across frames in preparation for task-specific processing. \Fig{fig:framework} shows an example that extracts temporal features across three frames: the current frame $F_t$ and the frame before ($F_{t-1}$) and after ($F_{t+1}$) the current frame, which we call the temporal frames. More temporal frames are possible in principle.

The front-end first calculates the flows between each temporal frame and the current frame using the algorithm described in \Sect{sec:design:motion}. A critical challenge we face is that the estimated flows are necessarily sparser than the corresponding image, because LiDARs generally have lower resolutions than that of cameras. For instance, the Velodyne HDL64E LiDAR, a high-end, high-density LiDAR, generates about 130,000 points per frame, whereas an image with a 720p resolution contains about 1 million points. \Fig{fig:comp}(a) illustrates the effect of using sparse point clouds, where only a small amount of pixels have points associated with them when projecting a single point cloud to the image.

To mitigate the sparsity of LiDAR-generated point clouds, we propose to register multiple point clouds together to form a dense point cloud. We register point clouds by simply transforming adjacent point clouds using the ego-motion $T_{ego}$ calculated from the IMU measurements (\Equ{equ:tran} and \Equ{equ:rot}). \Fig{fig:comp}(b) shows that when projecting multiple registered point clouds, many more pixels are associated with points.
%Alternatively, one could leverage computationally intensive, but potentially more accurate, registration algorithms (e.g., using Iterative Closest Point~\cite{besl1992method, chen1992object}).

Even with multiple point clouds, not every image pixel in $F_{t-1}$ (or $F_{t+1}$) has a corresponding flow. As a result, when warping images using flows the warped images will have many ``holes'', as illustrated in \Fig{fig:comp}(c). While one could merge more point clouds to increase the point density, doing so is susceptible to mis-registration, which is especially significant when merging a long sequence of point clouds where errors can accumulate.

To address this issue, we propose blocked warping, which duplicates a pixel's flow to its neighboring pixels (e.g., a $5\times5$ block) during warping. This is analogous to blocked-based motion compensation in conventional video compression. The assumption is that points corresponding to the neighboring pixels have similar motion in the 3D space, and thus their pixel flows are similar. We warp a temporal frame ($F_{t-1}$ or $F_{t+1}$) to the current frame using the blocked flows. The result is shown in \Fig{fig:comp}(d), which has much dense pixels (fewer ``holes'') than in \Fig{fig:comp}(c).

Finally, each warped temporal frame (e.g., $WF_{t-1}$), along with its unwarped counterpart (e.g., $F_{t-1}$) and the current frame ($F_{t}$), are concatenated and go through a convolutional layer to extract the temporal correlations between the temporal frame and the current frame. The features of the current frame are extracted independently.

\paragraph{Task-Specific Layers} The back-end of our architecture takes the extracted temporal features to perform video enhancement. The exact design of the back-end layers is task-specific. Our goal of this paper is \textit{not} to demonstrate new task-specific layers; rather, we show that our temporal feature extraction front-end is compatible with different task layers in a plug-and-play manner.

To that end, we implement three back-end designs for three video enhancement tasks, including super-resolution, denoising, and deblurring, by directly using designs from other algorithms (with slight modifications so that the interface matches our front-end). The layers for super-resolution and deblurring connect the temporal features from the front-end in a recurrent fashion, similar to designs of RBPN~\cite{haris2019rbpn} and ESTRNN~\cite{zhong2020estrnn}, respectively. The denoising layers concatenate the temporal features, which then enter a set of convolutional layers, similar to DVDnet~\cite{tassano2019dvdnet}.

%% file: setup.tex
\section{Evaluation Methodology}
\label{sec:exp}

\paragraph{Applications and Baselines} We evaluate three video enhancement tasks: super-resolution, deblurring and denoising.

\begin{itemize}
    \item Super-resolution: we compare with two DNN baselines: \mode{RBPN}~\cite{haris2019rbpn} %, EDVR~\cite{wang2019edvr}, 
    and \mode{VESPCN}~\cite{caballero2017vespcn}. \mode{RBPN} uses a recurrent encoder-decoder to learn temporal correlations; \mode{VESPCN} warps images in the pixel space and fuses multiple warped frames through a CNN to upsample.
    %EDVR is an implicit method that uses deformable convolution, 
    %Both two are explicit methods that require explicit pixel motion.
    \item Deblurring: we compare with \mode{ESTRNN}~\cite{zhong2020estrnn}, which uses RNN to learn the temporal features; we also compare with \mode{DeepGyro}~\cite{mustaniemi2019gyroscope}, which fuses IMU with image data for single-image deblur.
    \item Denoising: we compare with \mode{DVDnet}~\cite{tassano2019dvdnet}, which uses CNN to extract explicit motion and warp frames.
\end{itemize}

 In addition, we also designed a simple LiDAR-camera fusion baseline for \textit{each} task. This baseline, which we call \mode{VEFusion}, resembles many LiDAR/camera fusion DNNs~\cite{fu2019lidar}: it first concatenates the projected point cloud and the image; the concatenated data then enters the task-specific layers. Our proposed method also leverages point clouds for video enhancement, but uses point clouds in a different way: instead of fusing points with pixels, we use point clouds to generate flows. This baseline allows us to assess the effectiveness of this way of using point cloud for video enhancement. We make sure \mode{VEFusion} has roughly the same amount of parameters as our proposed method such that the performance difference is due to the algorithm.

%\paragraph{Implementation} Our framework allows us to simplify the baseline algorithms for faster execution speed. In particular, we made following simplifications to the baselines. For explicit method (RBPN, VESPCN, DVDnet), we replace their motion estimation module with our light-weight motion estimation module. For implicit method (ESTRNN), we remove its temporal alignment module (GSA), based on its implementation.

% the modifications are made based on their implementations. In EDVR, we remove the temporal alignment module (TSA) and reduce the number of deformation convolution layers in the PCD-Align module. In ESTRNN, we remove temporal alignment module (GSA).

\paragraph{Variants} We evaluate two variants of our methods: \mode{Ours-S} uses a single point cloud for flow estimation, and \mode{Ours-M} uses five point clouds for flow estimation.

% In this work, we choose two different ways to obtain camera transformation, one is using the IMU sensor, the other is a point cloud registration algorithm, we choose a ICP-based registration pipeline~\cite{xu2019tigris} developed using the widely-used PCL~\cite{rusu2011pcl}. To fully evaluate the effectiveness of our framework, we configure three variants:

%\begin{itemize}
%    \item Single point cloud with IMU (\mode{SPC+IMU}): this variant uses a single point cloud for motion estimation, and uses the IMU sensor to obtain the camera egomotion.
%    %\item Single point cloud with registration (\mode{SPC+R}): this variant uses a single point cloud for motion estimation, and uses point cloud registration to generate egomotion.
%    \item Multiple point clouds with IMU (\mode{MPC+IMU}): this variant uses IMU to obtain the camera egomotion, which is used to merge multiple point clouds to provide a denser motion estimation. This variant does \textit{not} use point cloud registration.
%\end{itemize}

%\paragraph{Training Settings} To train \mode{SPC+IMU}, we use the same training settings as the baselines and train on a single Nvidia RTX 2080 GPU. We then fine-tune \mode{SPC+IMU} to obtain \mode{MPC+IMU} and \mode{SPC+R} with 50 additional epochs \textit{each}. The training uses the Adam optimizer~\cite{kingma2014adam}.

\paragraph{Dataset} We use the KITTI dataset~\cite{Geiger2013kitti}, which provides sequences of synchronized LiDAR, camera, and IMU data. Following the common practices, we preprocess the dataset for different tasks. For super-resolution we downsize the videos by 4$\times$ in both dimensions using bicubic interpolation, similar to VESPCN~\cite{caballero2017vespcn}; for deblurring we add Gaussian blur to the videos, similar to EDVR~\cite{wang2019edvr}; for denoising we apply random noises to the videos, similar to DVDnet~\cite{tassano2019dvdnet}.

\paragraph{Evaluation Metrics} To evaluate the efficacy of our method, we use two metrics, PSNR and SSIM, to qualitatively evaluate the results. We also show the runtime performance of different methods by measuring the execution time of different methods on two platforms, one is the Nvidia RTX 2080 GPU; the other is the mobile Volta GPU on Nvidia's recent Jetson Xavier platform~\cite{xavier}. Each execution time is averaged over 1000 runs. 

\paragraph{Design Parameters} Unless otherwise noted, we use a block size of $3 \times 3$ in super resolution, and a block size of $7 \times 7$ in deblurring and denoising tasks. Five point clouds are registered for flow estimation. We will study the sensitivity to these two design parameters (\Sect{sec:eval:sen}).

%For all experimental evaluation, we use patch size of 3 in all super-resolution tasks, and patch size of 7 in video deblurring and denoising. In evaluation, we will further evaluate the sensitivity of patch size.
%
%Here, we choose to register 5 consecutive point clouds per image.

%% file: eval.tex
\section{Evaluation}
\label{sec:eval}

We show that the execution speed of our method is on average an order of magnitude faster than existing methods while at the same time delivering higher task quality, both objectively and subjectively (\Sect{sec:eval:overall}). We study the accuracy of our flow estimation (\Sect{sec:eval:flow}) and the sensitivity of our method on key design parameters (\Sect{sec:eval:sen}).

\subsection{Overall Evaluation}
\label{sec:eval:overall}

\paragraph{Results Overview} \mode{Ours-M} and \mode{Ours-S} consistently outperform the baselines in both quality and speed. \mode{Ours-M} is slightly better than \mode{Ours-S} due to the use of multiple point clouds for flow estimation. A naive fusion of point cloud and images, as done by \mode{VEFusion}, has significantly lower quality than our methods, albeit with a similar speed.

\begin{table}[ht]
\caption{Super-resolution comparison. Execution times are normalized to that on \mode{Ours-M}; H and M denote the high-end 2080 Ti GPU and the mobile Volta GPU, respectively.}
\centering
\resizebox{\columnwidth}{!}{
\renewcommand*{\arraystretch}{1}
\renewcommand*{\tabcolsep}{5pt}
\begin{tabular}{c c c c | c c} 
\toprule[0.15em]
 & \mode{RBPN} & \mode{VESPCN} & \mode{VEFusion} & \mode{Ours-S} & \mode{Ours-M}  \\
\midrule[0.05em]
PSNR (dB) & 27.08 & 24.78 & 26.95 & 27.43 & \textbf{27.50} \\ 
SSIM & 0.860 & 0.787 & 0.854 & \textbf{0.873} & 0.872 \\
\midrule[0.05em]
Time (H) & 36.10 & 0.55 & 1.00 & 1.00 & 1.00 \\ 
Time (M) & 7.24 & 0.13 & 1.00 & 1.00 & 1.00 \\
\bottomrule[0.15em]
\end{tabular}
}
\label{table:super_res}
\end{table}

\paragraph{Super-resolution} 
\Tbl{table:super_res} compares different super-resolution algorithms. We also show the execution time of different methods normalized to that of \mode{Ours-M}.

Overall, \mode{Ours-M} achieves the highest visual quality both in terms of PSNR and SSIM among all methods. \mode{Ours-S} has similar SSIM but lower PSNR. \mode{Ours-M} achieves a 36.10$\times$ speedup against \mode{RBPN} on 2080 Ti and 7.24$\times$ speedup on the mobile GPU, showing the effectiveness of our lightweight flow estimation algorithm, which executes in about $10 \mu s$ on GPUs. \mode{Ours-M} and \mode{Ours-S} have virtually the same speed, because transforming point clouds into one frame has negligible overhead. \mode{VEFusion} has the same speed as our methods with lower quality. \mode{VESPCN} is the fastest, but has a much lower super-resolution quality due to a simpler CNN.

\begin{table}[ht]
\caption{Deblurring comparison.}
\centering
\resizebox{\columnwidth}{!}{
\renewcommand*{\arraystretch}{1}
\renewcommand*{\tabcolsep}{5pt}
\begin{tabular}{c c c c | c c} 
\toprule[0.15em]
 & \mode{ESTRNN} & \mode{DeepGyro} & \mode{VEFusion} & \mode{Ours-S} & \mode{Ours-M}  \\
\midrule[0.05em]
PSNR (dB) & 34.78 & 31.20 & 35.22 & 35.50 & \textbf{36.61} \\ 
SSIM & 0.945 & 0.806 & 0.949 & 0.950 & \textbf{0.957} \\
\midrule[0.05em]
Time (H) & 1.78 & 6.20 & 1.00  & 1.00 & 1.00 \\
Time (M) & 1.08 & 11.96 & 1.00  & 1.00 & 1.00 \\
\bottomrule[0.15em]
\end{tabular}
}
\label{table:deblur}
\end{table}

\paragraph{Deblurring} \Tbl{table:deblur} compares different methods on video deblurring. Our method, \mode{Ours-M}, achieves the highest quality both in terms of PSNR and SSIM. Compared to \mode{ESTRNN}, \mode{Ours-M} achieves 1.83 higher in PSNR and 0.012 higher in SSIM. Our methods are also faster than the baselines on both GPUs. The speedup on \mode{ESTRNN} is not significant, because the flow estimation in \mode{ESTRNN} is small to begin with (7.7\% on the mobile GPU). \mode{DeepGyro} has the lowest task quality and the slowest speed. Its low quality is mainly attributed to the fact that it deblurs using a single image, while other methods use temporal information.

\begin{table}[ht]
\caption{Denoising comparison.}
\centering
\resizebox{\columnwidth}{!}{
\renewcommand*{\arraystretch}{1}
\renewcommand*{\tabcolsep}{10pt}
\begin{tabular}{c c c | c c} 
\toprule[0.15em]
 & \mode{DVDnet} & \mode{VEFusion} & \mode{Ours-S} & \mode{Ours-M}  \\
\midrule[0.05em]
PSNR (dB) & 27.19 &  31.60 & 33.34 & \textbf{33.89} \\ 
SSIM & 0.838 & 0.951 & 0.953 &  \textbf{0.961} \\
\midrule[0.05em]
Time (H) & 187.41 & 1.00  & 0.99 & 1.00 \\
Time (M) & 68.97 & 1.00  & 0.99 & 1.00 \\
\bottomrule[0.15em]
\end{tabular}
}
\label{table:denoise}
\end{table}

\paragraph{Denoising} For video denoising, \mode{Ours-M} achieves the highest quality both in PSNR and SSIM, as shown in \Tbl{table:denoise}. \mode{Ours-M} improves upon \mode{VEFusion} and \mode{DVDnet} by a large margin --- 2.29 dB and 6.70 dB in PSNR, respectively. Meanwhile, \mode{Ours-M} has a 187.4$\times$ speedup compared to \mode{DVDnet} on 2080 Ti and 69.0$\times$ speedup on the mobile GPU. The speedup comes from avoiding the expensive flow estimation algorithm DeepFlow~\cite{weinzaepfel2013deepflow} used in \mode{DVDnet}.

\begin{figure*}[t!]
  \centering
  \includegraphics[width=\textwidth]{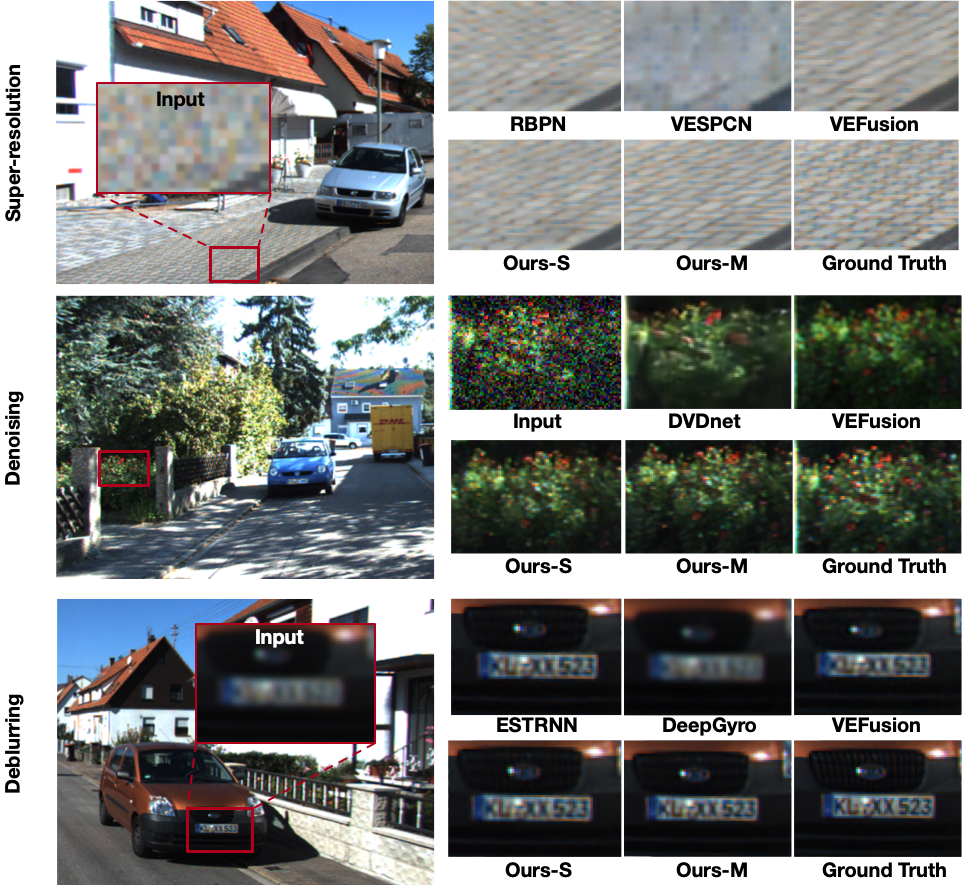}
  %\vspace{-15pt}
  \caption{Visual comparison of different methods on various visual enhancement tasks.}
  \label{fig:examples}
\end{figure*}

% \paragraph{Objective Comparison} Our framework consistently outperforms the baselines in both PSNR and SSIM. \Tbl{table:psnr_ssim} shows the performance comparison between the three variants of our method and the corresponding baselines. SPC+IMU, our fastest variant, has higher task quality than all the baselines. SPC+R generally has higher quality than SPC+IMU, because the former uses registration to generate a more accurate camera egomotion (compared to using the IMU).

% MPC+IMU registers multiple point clouds for motion estimation, and has the highest quality for almost all tasks. We attribute this to the effectiveness of a denser point cloud in our motion estimation. With the emergence of high-resolution 3D point cloud acquisition devices~\cite{scarlet, zed2}, we expect that our framework can readily leverage these devices to provide higher quality video enhancement.

\paragraph{Subjective Comparison} Our approach is also visually better than the baselines upon subjective comparisons. \Fig{fig:examples} shows the visual comparisons on different tasks. The improvements from the baselines to \mode{Ours-M} are the most significant. \mode{Ours-M} is best at revealing details, such as the roads and bushes, because of its dense motion obtained from merging point clouds.

%This result demonstrates the effectiveness of our method, without proposing any new DNN algorithm, our method can reduce computation in DNN algorithms and achieve better visual quality than baselines.

\subsection{Flow Estimation Accuracy and Speed}
\label{sec:eval:flow}

Our lightweight flow estimation algorithm provides accurate flow information. To demonstrate the effectiveness of the estimated flows, we warp frames in the dataset using the estimated flows and calculate the PSNR. \Tbl{table:flow} shows the results across different flow estimation algorithms used in different networks. We also show the speed of different flow estimation algorithms normalized to that of ours.

\begin{table}[ht]
\caption{Flow estimation comparison. Execution time is normalized to that of ours.}
\centering
\resizebox{\columnwidth}{!}{
\renewcommand*{\arraystretch}{1}
\renewcommand*{\tabcolsep}{10pt}
\begin{tabular}{c c c c | c} 
\toprule[0.15em]
 & \mode{DVDnet} & \mode{VESPCN} & \mode{RBPN} & \mode{Ours}  \\
\midrule[0.05em]
PSNR (dB) & 14.71 & 16.64 & 22.68 & 18.74 \\ 
\midrule[0.05em]
Time (H) & 4147.5 & 1420.0 & 98694.0 & 1.0 \\ 
\bottomrule[0.15em]
\end{tabular}
}
\label{table:flow}
\end{table}

Judged by the quality of warped images, our flow estimation method is better than the estimation methods used in \mode{DVDnet} and \mode{VESPCN}, as shown in \Tbl{table:flow}. This also explains the task quality difference. Interestingly, while the frames warped using our flow estimation have a lower PSNR compared to those in \mode{RBPN}, we are able to achieve a better super-resolution quality than \mode{RBPN}. The reason is that our method uses the warped frames to extract temporal features (\Fig{fig:framework}) while \mode{RBPN} uses the actual flow values.

Our flow estimation is also at least three orders of magnitude faster than other methods used in baselines. This explains the overall speed difference shown earlier, since our task-specific layers are similar to those used in the baselines.

\subsection{Sensitivity Study}
\label{sec:eval:sen}

We use super-resolution as an example to study how the block size used in blocked warping and the number of merged point clouds used in flow estimation influence the task quality. Other tasks have a similar trend.

\begin{table}[ht]
\caption{Sensitivity of the block size and the number of merged point clouds on super-resolution on \mode{Ours-S}.}
\centering
\resizebox{\columnwidth}{!}{
\renewcommand*{\arraystretch}{1}
\renewcommand*{\tabcolsep}{10pt}
\begin{tabular}{c c c c c}
\toprule[0.15em]
Patch size & $1\times 1$ & $3\times 3$  & $5\times 5$ & $7\times 7$  \\
\midrule[0.05em]
PSNR (dB) & 27.02 & \textbf{27.43} & 27.29 & 27.26 \\ 
\bottomrule[0.15em]
\\[-1.5ex]
%\toprule[0.15em]
 \# of point clouds & 1 & 3  & 5 & 7 \\
\midrule[0.05em]
PSNR (dB) & 27.43 & 27.47 & 27.50 & \textbf{27.52}  \\ 
\bottomrule[0.15em]
\end{tabular}
}
\label{table:sen}
\end{table}

\paragraph{Block Size} Larger blocks initially improve the task quality. \Tbl{table:sen} shows how the super-resolution quality varies with the block size. When the block size initially increases from $1 \times 1$ to $3 \times 3$, the PSNR improves because the flow density increases. Increasing the block size further degrades the quality. This is because with large blocks more pixels'  flows are duplicated from neighbor pixels rather than calculated using depth information, reducing the flow accuracy.

\paragraph{Number of Merged Point Clouds} Merging more point clouds leads to denser and more accurate flow estimation and thus a higher the task quality. This is evident in \Tbl{table:sen}, which shows that the PSNR of increases as the number of merged point clouds increases.

%We note that merging more point clouds has little speed overhead since we merge point clouds using the egomotion calculated from IMU measurements (\Equ{equ:tran} and \Equ{equ:rot}), which executes in a few microseconds as discussed before. Our current implementation chooses to merge 5 point clouds.

%It does mean that more point cloud data needs to be stored and transmitted. However, we expect that the storage overhead will become smaller in the future as point cloud compression techniques become more mature~\cite{feng2020real, vpcc2019, gpcc2019}.

%% file: conc.tex
\section{Conclusion}
\label{sec:conc}

We demonstrate a general framework to build fast and accurate video enhancement algorithms. The key is to assist video enhancement with an accurate depth-driven flow estimation algorithm. Our flow estimation is accurate because it leverages the accurate depth information generated from LiDARs based on  a physically-plausible scene model. We show strategies to overcome the sparsity of LiDAR point clouds. Our flow estimation is lightweight because it relies on only simple geometric transformations, enabling lean end-to-end algorithms. We propose a generic framework that integrates the flow estimation with task-specific layers in a plug-and-play manner. We achieve over an order of magnitude speedup while improving task quality over competing methods. While fusing point clouds with images has been extensively studied lately in vision tasks, we show that using point clouds for flow estimation, rather than simply fusing them with images, achieves better performance.

An implication of our framework is that the point cloud data must be attached to the video content, which could potentially increase the storage and transmission overhead. However, the overhead is likely small, because the size of point cloud data is smaller than that of images. For instance, one point cloud frame obtained from a high-end Velodyne HDL-64E LiDAR~\cite{VelodyneHDL64E} is about 1.5 MB, whereas one 1080p image is about 6.0 MB in size. The overhead will become even smaller in the future as point cloud compression techniques become more mature~\cite{feng2020real, vpcc2019, gpcc2019}.

%The point cloud frame rate from LiDAR is lower than camera, but if the point cloud is from depth cameras, then the rates will match.